\documentclass[letterpaper, 10 pt, conference]{ieeeconf}  % Comment this line out if you need a4paper

\IEEEoverridecommandlockouts                              % This command is only needed if 
\usepackage{amsmath} % assumes amsmath package installed
\usepackage{amssymb}  % assumes amsmath package installed
\usepackage{cite}
\usepackage[ruled]{algorithm2e}
\usepackage{graphicx}
\usepackage{textcomp}
\usepackage{xcolor}
\DeclareMathOperator*{\argmax}{arg\,max}
\usepackage{booktabs,caption}
\usepackage[flushleft]{threeparttable}

\title{\LARGE \bf
Safe Decision-making for Lane-change of Autonomous Vehicles via Human Demonstration-aided Reinforcement Learning
}

\author{Jingda Wu$^{1}$, Wenhui Huang$^{1}$, Niels de Boer$^{2}$, Yanghui Mo$^{2}$, Xiangkun He$^{1}$ and Chen Lv$^{1}$% <-this % stops a space
\thanks{$^{1}$Jingda Wu, Wenhui Huang, Xiangkun He, and Chen Lv are with School of Mechanical and Aerospace Engineering, Nanyang Technological University, 50 Nanyang Ave, Singapore}%
\thanks{$^{2}$Niels de Boer and Yanghui Mo are with the CETRAN AV Testing Center, Energy Research Institute, Nanyang Technological University, 50 Nanyang Ave, Singapore}%
\thanks{Corresponding author: Chen Lv. (E-mail: lyuchen@ntu.edu.sg)}
}

\begin{document}

\maketitle
\thispagestyle{empty}
\pagestyle{empty}

%%%%%%%%%%%%%%%%%%%%%%%%%%%%%%%%%%%%%%%%%%%%%%%%%%%%%%%%%%%%%%%%%%%%%%%%%%%%%%%%
\begin{abstract}

Decision-making is critical for lane change in autonomous driving. Reinforcement learning (RL) algorithms aim to identify the values of behaviors in various situations and thus they become a promising pathway to address the decision-making problem. However, poor runtime safety hinders RL-based decision-making strategies from complex driving tasks in practice. To address this problem, human demonstrations are incorporated into the RL-based decision-making strategy in this paper. Decisions made by human subjects in a driving simulator are treated as safe demonstrations, which are stored into the replay buffer and then utilized to enhance the training process of RL.  A complex lane change task in an off-ramp scenario is established to examine the performance of the developed strategy. Simulation results suggest that human demonstrations can effectively improve the safety of decisions of RL. And the proposed strategy surpasses other existing learning-based decision-making strategies with respect to multiple driving performances.

\end{abstract}

%%%%%%%%%%%%%%%%%%%%%%%%%%%%%%%%%%%%%%%%%%%%%%%%%%%%%%%%%%%%%%%%%%%%%%%%%%%%%%%%
\section{INTRODUCTION}

The decision-making function that receives the ambient environment information and generates high-level intentions for autonomous vehicles (AVs) is a crucial component in devising the driving strategy \cite{yang_jas,cognitive_design,cyber_attack,av_survey}. The early decision-making strategies were based on rules, but they were not adequate to cover all scenarios. As deep learning technology reaches maturity, deep reinforcement learning (RL), which show great representability and optimization ability, are promising for developing decision-making strategies for automated vehicles \cite{wu2022uncertainty}. 

A majority of mainstream RLs are value-iteration algorithms, which target the establishment of a value approximator and make sure that optimal values are achieved at all times \cite{sutton2018reinforcement}. Far-reaching value-iteration RLs such as Deep Q Network (DQN) have been applied to decision makings of AVs in city roads \cite{chen2020conditional} and highways \cite{wang2021interpretable}. In \cite{friji2020dqn}, camera images were used to construct the state space for a DQN-based car-following strategy. In \cite{zhang2021tactical}, the historic visual information was employed to construct a recurrent DQN algorithm for tackling the behavior decision making problem. However, the strategies based on value-iteration RLs suffer from poor safety. Specifically, RLs are likely to take actions that cause AV to collide with the surrounding environment. This phenomenon is attributed to the algorithm nature: value-iteration RLs have to visit numerous high-value data before they can learn to perform safe behaviors. However, the random exploration mechanism of RL is likely to cause insufficient high-quality data, whereby the RL-based strategy exhibits inferior safety.

Human demonstrations, as prior knowledge, are expected to mitigate the above problem and improve the safety of decisions \cite{wu2021prioritized}\cite{huang2022efficient}. The value approximator trained with safe human demonstrations can learn the value of favorable behaviors and thus avoid catastrophic actions. In \cite{rajeswaran2017learning}, human demonstrations for manipulating the robotic arm were collected beforehand and utilized to improve RL in controlling the robots safely. In \cite{wang2018intervention}, human data in operating the drone was used for augmenting the RL-driven controller with safety requirements. In \cite{wu2020battery}, a rule-based system generated expert demonstrations in the training session of RL to enhance the performance of the energy management strategy. In \cite{gulcehre2019making}, dual experience replay buffers were adopted to store the human demonstrations and exploration data of RL, respectively. As a consequence, the DQN algorithm addressed six games safely that had never been overcome by similar algorithms. Despite the above efforts, there are few studies attempting to introduce demonstrations for RL-based safe decision-making strategies of AVs.

\begin{figure*}[htp]
    \centering
    \includegraphics[width=\linewidth]{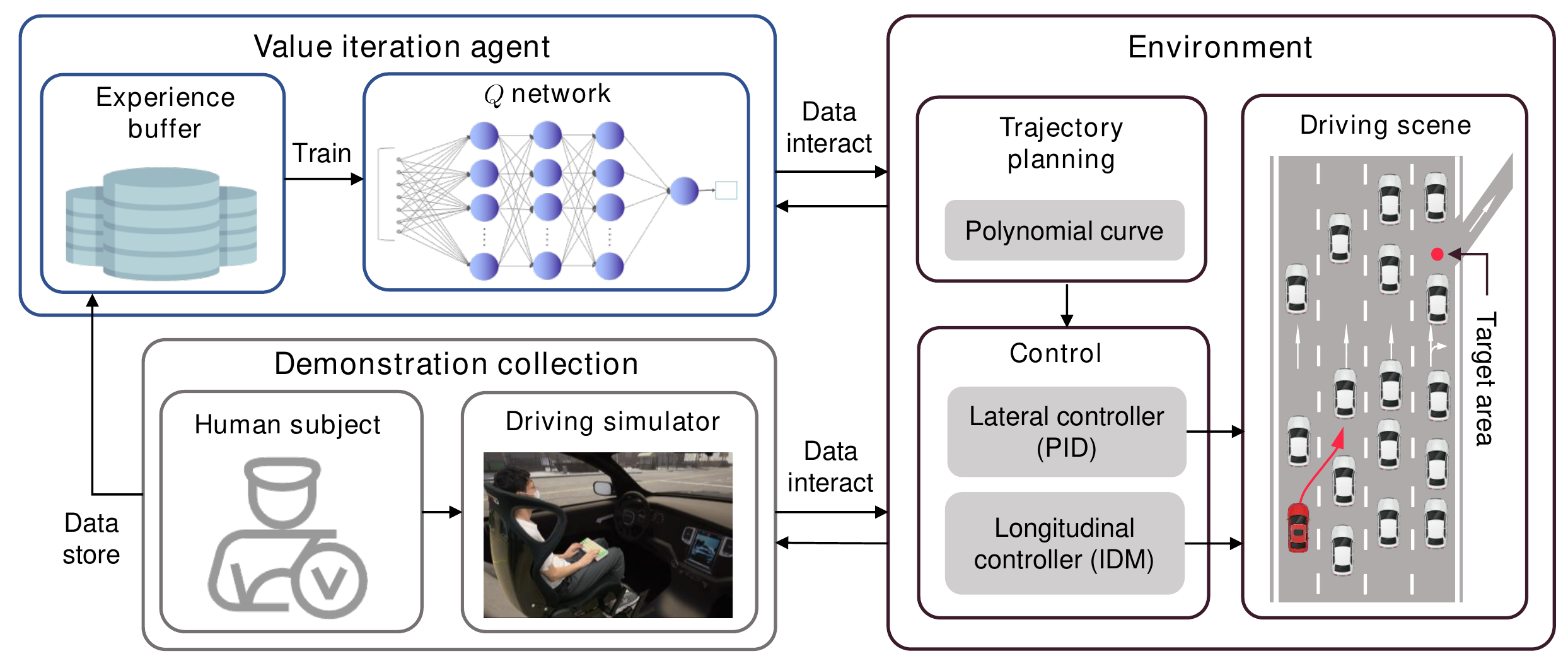}
    \caption{Schematic diagram of the proposed method.}
    \label{fig1}
\end{figure*}

This paper aims to bridge the abovementioned gap. A human demonstration-aided value iteration method is proposed to improve the safety of deep RL in establishing the decision-making strategy. The schematic diagram of the proposed method is illustrated in Figure \ref{fig1}. Double Dueling DQN (D3QN)\cite{wang2016dueling}, a state-of-the-art value iteration-based algorithm, was employed as the backbone to conduct the decision makings. We enabled proficient human subjects to execute the driving task by a driving simulator and collect the generated data. Then, the data, which can be viewed as safe demonstrations, was sent to the experience replay buffer of D3QN. The safety, efficiency, and asymptotic performance of the D3QN agent were expected to be improved through the aid of safe demonstrations. The proposed strategy was examined through a lane change scenario, wherein the strategy generated high-level behavioral decisions for guiding the ego vehicle. We adopt two learning-based strategies as baselines to evaluate the ability of algorithms: a imitation learning-based strategy, which was also trained by human demonstrations; and a strategy based on the vanilla D3QN algorithm, which is ablated from human demonstrations. 

The main contributions of this research are summarized as follows: 1) a human demonstration-aided RL approach is proposed, wherein the safety and performance of value-iteration-based RL algorithms are improved. 2) a safe decision-making strategy based on the proposed framework is established and examined in a lane change task, and its ability in controlling AVs is evaluated through comparison with existing learning-based strategies.

The remainder of the paper is organized as follows. The proposed method is elaborated on in Section II. The problem formulation of lane change is described in Section III. The implementation details of data collection and experiments, and the validation results are provided in Section IV. Finally, conclusions are drawn in Section V.

\section{Proposed Method}\label{section2}
In the proposed method, human demonstrations are utilized to improve the safety and performance of value-based RL algorithms. First, the D3QN algorithm, which serves as the basis for the proposed value-based RL, is introduced in principle. Then, human demonstrations are incorporated into the D3QN algorithm to form the proposed method.

\subsection{Double Dueling Deep Q Network}\label{section2.1}

Double Dueling Deep Q Network (D3QN) is an advanced variant of the DQN algorithm to address the overestimation problem and improve sampling efficiency. The interaction process between D3QN and the controlled environment is described using a Markov Decision Process (MDP). At a time-step \(t\), the D3QN agent sends an action \(a_t\) to the environment. The sate transitions \(s_t \rightarrow s_{t+1}\) and reward signal \(r_t\), as feedbacks to the action, are generated in the environment. The D3QN algorithm establishes a value function \(Q\) to calculate the cumulative reward $\sum_{t}^{\infty} \gamma^{t}\cdot r_t$ under any pair of \((s_t,a_t)\), where \(\gamma\in(0,1]\) is the discount factor. The above process is formulated using the Bellman equation, as: 
\begin{equation}
Q\left(s_t,a_t\right)=r_t+\gamma\cdot\mathbb{E}\left[Q\left(s_{t+1},a_{t+1}\right)\right]
\end{equation}

The D3QN algorithm aims to obtain the optimal policy \(\pi^\ast\), which executes the actions leading to the maximum \(Q\) value at any time step, represented as:
\begin{equation}
    \pi^\ast\left(s_{t}\right) = \argmax_a Q(s_{t},a)
\end{equation}

Then the objective turns to find an accurate value function \(Q\). In this regard, a neural network with parameter \(\theta\) is used to approximate the value function. Hence, the value function can be represented by \(Q\left(s,a\middle|\theta\right)\) network, and the policy can be represented as \(\pi\left(s\middle|\theta\right)\). Specific to D3QN, two same-structure \(Q\) networks are used to address the over-estimation problem. The loss function \(\mathcal{L}\) of two \(Q\) network are calculated as:
\begin{subequations}
\begin{align}
        \mathcal{L}_1=\mathbb{E}\left[r_t+\gamma Q_2\left(s_{t+1},\pi\left(s_{t+1}\right)\middle|\theta_2\right)-Q_1\left(s_t,a_t\middle|\theta_1\right)\right]\\
        \mathcal{L}_2=\mathbb{E}\left[r_t+\gamma Q_1\left(s_{t+1},\pi\left(s_{t+1}\right)\middle|\theta_1\right)-Q_2\left(s_t,a_t\middle|\theta_2\right)\right]
\end{align}
\end{subequations}
where \(\theta_1\) and \(\theta_2\) denote the parameters of two \(Q\) networks, respectively.

In D3QN, the dueling mechanism is utilized to improve the sample efficiency. Specifically, the body of \(Q\) network is split into two branches, which are the state-value network \(V(s)\) and advantage network \(A(s, a)\), respectively. This mechanism can be represented by:
\begin{equation}
    Q\left(s_t,a_t\middle|\theta\right)=V\left(s_t|\theta^{br1}\right)+A\left(s_t,a_t\middle|\theta^{br2}\right)
\end{equation}
where \(\theta^{br1}\) and \(\theta^{br2}\) denote the parameters of two branches, respectively.

The data utilized in the above process in one time step is a tuple \(\xi\left(\cdot\right)\) which contains four elements, and the data is stored to an experience buffer \(\mathcal{B}\) for replay. This process is represented as:
\begin{equation}
    \mathcal{B}\gets\mathcal{B}\cup\xi_t\left(s_t,a_t,r_t,s_{t+1}\right)
\end{equation}

\subsection{Human-demonstration-aided Algorithm}\label{section2.2}
The learning performance of D3QN is heavily related to sample quality. In the conventional D3QN, sample data is obtained through the \(\varepsilon-\)greedy method. During the interaction between the agent and the controlled environment, the sample tuple is obtained by:
\begin{equation}
    a_t=\ \varepsilon\cdot(a \sim U)+\left(1-\varepsilon\right)\cdot\pi\left(s_t\right|\theta)
\end{equation}
where \(\varepsilon\in\left[0,1\right]\) is the greedy variable that varies with the training process, and \(U\) represents uniform distribution.

The above random sampling is inefficient owing to the lack of prior knowledge. Human demonstrations are added to the sample distribution; this process could enable D3QN to visit favorable actions more frequently and learn the optimal policy with a higher efficiency \cite{wu2021human}. 

Human demonstrations are collected before the learning process of D3QN, and the generated tuple \(\xi^\mathcal{H}\) is shown as:
\begin{equation}
    \xi^\mathcal{H}=\left(s_t,a_t^\mathcal{H},r_t,s_{t+1}\right)
\end{equation}
where \(a^\mathcal{H}\) represents the human demonstration action.

D3QN is expected to learn from a mixture of conventional experience data and human demonstration data. Supposing a data batch contains \(n_1\) human-demonstration data and \(n_2\) conventional data, the loss function of the \(Q_1\) network can be represented as:
\begin{equation}
\begin{split}
        \mathcal{L}_1=\frac{\sum_i^{n_1}|| r_i+\gamma \cdot Q_2 (s_{i+1},\pi(s_{i+1} )| \theta_2 )-Q_1 (s_i,a_i^\mathcal{H} |\theta_1 )||_2^2}{n_1}  \\+ \frac{\sum_i^{n_2}|| r_i+\gamma \cdot Q_2 (s_{i+1},\pi(s_{i+1} )| \theta_1 )-Q_1 (s_i,a_i |\theta_2 )||_2^2 }{n_2}
\end{split}
\end{equation}

The process of the \(Q_2\) network is likewise calculated.

Lumping the above factors, the procedure of the proposed human-demonstration-aided D3QN algorithm is provided in Algorithm 1. Note that we use a single network \(Q\) to denote the value function for brevity as the double-\(Q\) setting is not the focus of the study.

\begin{algorithm}
\caption{Our algorithm}\label{alg1}
 Initialize \(Q\) network parameter \(\theta\)\;
 \For{episode=1 to \(E_1\)}{
  Initialize the environment \(s_t\sim Env\)\;
  \While{not done}{
    Record human action \(a_t^\mathcal{H}\)\;
    Interact with the environment \(r_t,s_{t+1}\sim Env\)\;
    Store the tuple \(\mathcal{B}\gets\mathcal{B}\cup\left(s_t,a_t^\mathcal{H},r_t,s_{t+1}\right)\)\;
  }
 }
 \For{episode=1 to \(E_2\)}{
 Initialize the environment \(s_t\sim Env\)\;
 \While{not done}{
 Sample action \(a_t\) from using \(\varepsilon\)-greedy policy\;
 Interact with the environment \(r_t,s_{t+1}\sim Env\)\;
 Store the tuple \(\mathcal{B}\gets\mathcal{B}\cup\left(s_t,a_t,r_t,s_{t+1}\right)\)\;
 Sample a batch of the data \(\left(s_i,a_i,r_i,s_{i+1}\right)\sim \mathcal{B}\)\;
 Calculate loss function \(\mathcal{L}\left(\theta\right)\) using Eq (8)\;
 Update \(Q\) network parameter \(\theta\)\;
 }
 }
\end{algorithm}

\section{Problem Formulation}\label{section3}
The problem to be addressed by the proposed algorithm is a lane change task. First, scenario details are described; then the decision-making problem is defined in the context of RL. 

\subsection{Scenario Description}\label{section3.1}
The lane change task occurs on a four-lane city road and the ego vehicle tends to exit the road. This requires the ego vehicle to maneuver from the leftmost lane to the shoulder lane. There were 15 vehicles surrounding the ego vehicle, whose positions and target speeds were randomly determined. The longitudinal behaviors of surrounding vehicles were controlled by the intelligent driver model (IDM), and their lane-change behaviors were generated and controlled by the built-in traffic manager of the simulation software. All vehicles travel along the center of the lane and travel at a velocity of 20 to 50 km/h. If the ego vehicle cannot drive the shoulder lane within 240 meters from its starting point, it is considered to have failed the test.

The RL outputs behavioral decision-making commands as its actions. Motion-planning and tracking functions convert decisions into specific longitudinal and lateral signals for vehicle control. Specifically, the polynomial planning method is used to generate the trajectory, wherein a proportional-integral-derivative controller (PID) and an intelligent driver model (IDM) conduct tracking controls. 

\begin{figure}[htp]
    \centering
    \includegraphics[width=\linewidth]{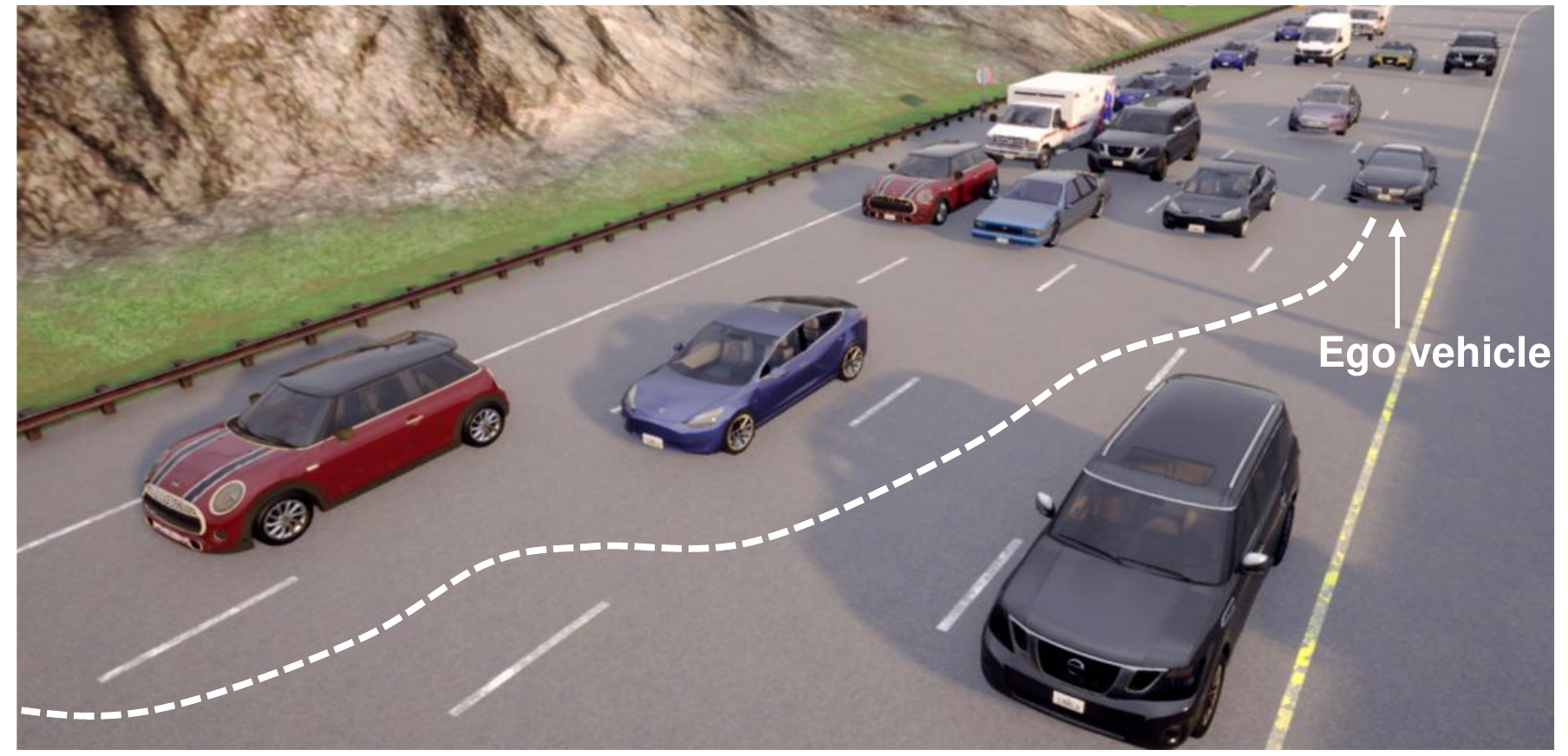}
    \caption{Illustration of the scenario. Ego conducts lane changes, and the dotted line indicates one possible route.}
    \label{fig2}
\end{figure}

The scenario is established using the CARLA simulator \cite{dosovitskiy2017carla}, and the scenario overview is shown in Figure \ref{fig2}.

\subsection{RL definition}\label{section3.2}
\textit{State variable}. The bird’s-eye-view image is chosen as the state variable to comprehensively describe the environment information. The images are collected per 0.5s and four consecutive images consist of a state variable. The width and height of the image are 80 and 45 pixels, respectively. 

\textit{Action variable}. Five actions: maintaining, accelerating, braking, leftward lane-change, and rightward lane-change constitute the action space. Specifically, A maintaining command keeps the current speed and lane, the accelerating and braking signal enable the ego vehicle to respectively increase and decrease the target velocity by 2 km/h, Two lane-change commands refer to changing to the target lane with an invariant cruise speed. The action is sent to the environment per 0.5s.

\textit{Reward function}. The reward function encourages the ego vehicle to conduct rightward lane-change behaviors for exiting the road, and the process should be safe and smooth as much as possible. Specifically, the reward function is calculated as:
\begin{subequations}
\begin{align}
    r_{t,1}&=s_t\in\mathcal{S}_{R}\\
    r_{t,2}&=s_t\in\mathcal{S}_{C}\\
    r_{t,3}&=\left[TTC\left(ego,front\right)\right]^{-1}\cdot\left[front\notin\emptyset\right]\\
    r_{t,4}&=\left[TTC\left(ego,rear\right)\right]^{-1}\cdot\left[rear\notin\emptyset\right]\\
    r_t&=r_{t,1}+r_{t,2}+r_{t,3}+r_{t,4}
\end{align}
\end{subequations}
where \(\mathcal{S}_{R}\) and \(\mathcal{S}_{C}\) denotes the state of rightward lane-change and collision, respectively, \(TTC\) denotes the time-to-collision metric between two vehicles, \(ego\), \(front\), and \(rear\) refer to the abbreviation of the ego vehicle, front surrounding vehicle and rear surrounding vehicle.

\textit{Neural network detail and hyperparameter}. The proposed algorithm is backboned on the autoencoder neural network, of which the structure is provided in Table \ref{table1}. The used hyperparameters of the RL algorithm are listed in Table \ref{table2}.

\begin{table}[htbp]
\caption{Neural Network Structure and Parameters}
\begin{center}
\begin{tabular}{ cc } 
 \hline
 \textbf{Parameters} & \textbf{Value} \\ 
 \hline
 Input Image shape & [80,45,4] \\ 
 Convolution Filter Features & [16,32,64] (kernel size 3\(\times\)3), stride=2 \\ 
 Fully Connected Layer Features & [256,128,5] \\
 \hline
\label{table1}
\end{tabular}
\end{center}
\end{table}

\begin{table}[htbp]
\caption{Hyperparameters of RL Algorithm}
\begin{center}
\begin{tabular}{ ccc } 
 \hline
 \textbf{Parameters} & \textbf{Description} & \textbf{Value} \\ 
 \hline
 Buffer size & Capacity of the experience replay buffer & 1e6 \\ 
 Max episode &Cutoff episode of the training process & 1000 \\ 
 Minibatch size & Capacity of minibatch & 64 \\
 Learning rate & Initial learning rate of the \(Q\) network & 0.005\\
 Activation & Activation method of layers of the network & relu\\
 Init exploration & Initial exploration rate in \(\epsilon\)-greedy & 1\\
 Cut exploration & Cutoff exploration rate in \(\epsilon\)-greedy & 0.1\\
 Gamma & Discount factor of the \(Q\) function & 0.9\\
 \hline
\label{table2}
\end{tabular}
\end{center}
\end{table}

\section{Experiment}\label{section3}
In this section, the implementation details of experiments are first provided. Then, the experimental results are shown and analyzed to evaluate the proposed strategy.

\subsection{Environment Configuration}\label{section4.1}

The scenario of Section \ref{section3.1} was implemented using the Town04 map of the CARLA simulator. The algorithm and related control procedures were programmed in python, and the neural networks were created using PyTorch. When collecting demonstrations, a human subject, who is proficient in driving and familiar with the control interface, was employed to execute the lane change task by pressing keys on the keyboard. Specifically,  “W-S-A-D” corresponded to behavioral decisions accelerating, braking, leftward lane-change, and rightward lane-change, respectively. And "space" corresponded to the maintaining command. A human action that was treated as a safe demonstration was broadcast to the environment every 0.5 seconds in line with the RL control period. In contrast to decision-making actions that respond per 0.5s, the tracking control modules respond per 0.02s. This allows the low-level functions have ample time to respond to high-level decisions.

Two baseline strategies for decision-making were adopted to evaluate the performance of the proposed algorithm. The first baseline is based on the vanilla D3QN, which does not use any human demonstrations in the training process. The other candidate is the imitation learning (IL)-based strategy, which utilizes only the human demonstrations to train the neural network with the behavior cloning objective.

\subsection{Results and Analysis}\label{section4.2}
The proposed strategy was evaluated in two aspects, i.e., training performance, and decision-making ability. 

\begin{figure}[htp]
    \centering
    \includegraphics[width=8cm]{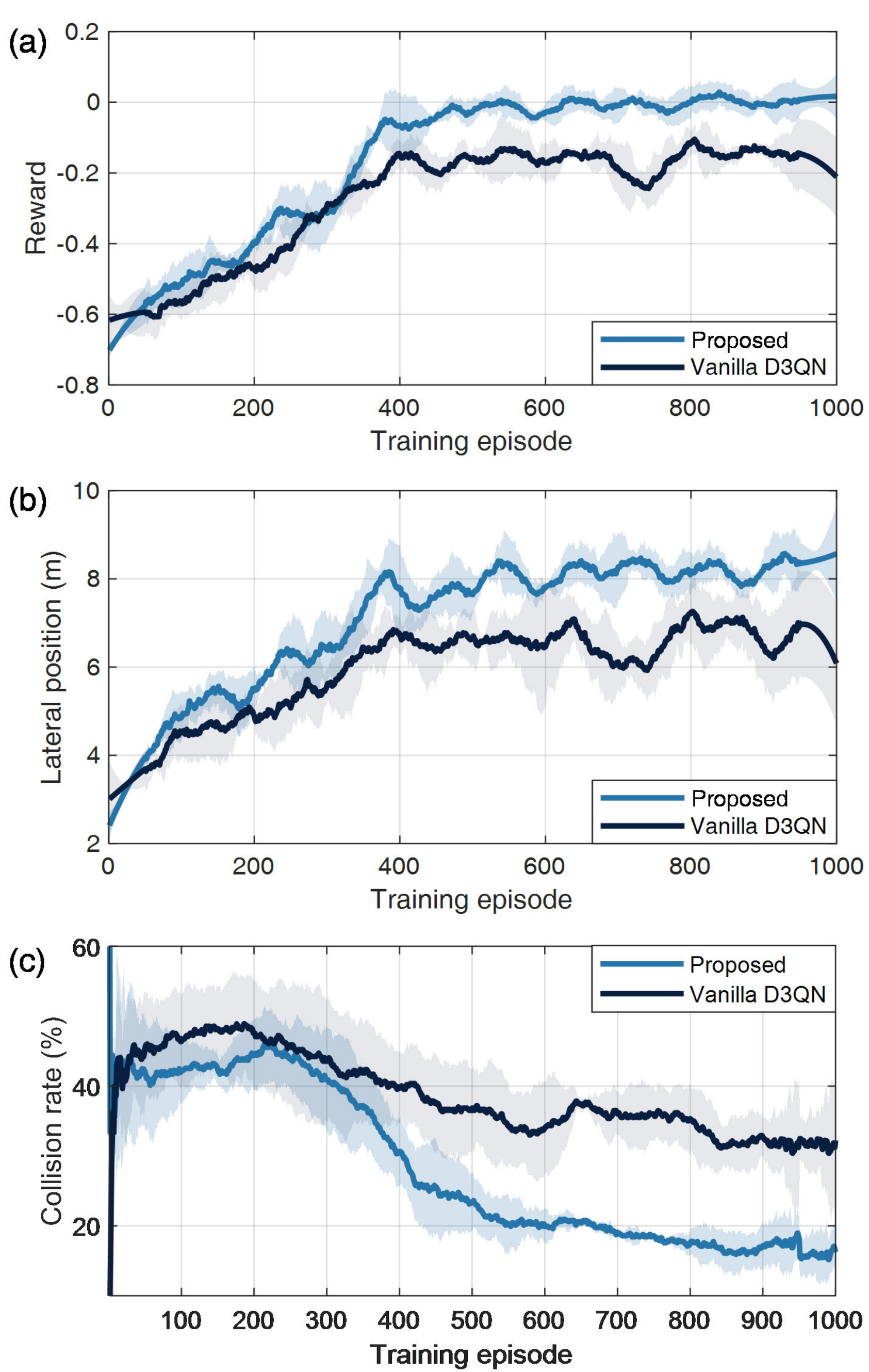}
    \caption{Training results of the proposed strategy and the vanilla D3QN-based strategy. Smoothed curves of (a) the episode reward in the training process, (b) the lateral position of the ego vehicle at the end of the episode in the training process, wherein the lateral position indicates the target achievement of the reward function, and (c) the collision rate of the ego vehicle in the training process.}
    \label{fig3}
\end{figure}

The training sessions of the proposed and vanilla-D3QN-based strategies were compared to evaluate their performance. Five training attempts under different random seeds were conducted for each strategy, and results were represented by a solid line of the mean value and an error bar of the standard deviation. Two metrics: reward and lateral position, were used to quantify the training results. The curves are illustrated in Figure  \ref{fig3}. As shown in Figure \ref{fig3}(a), the proposed strategy exhibits a faster learning speed, and the asymptotic reward of the proposed strategy was significantly higher than that of the baseline. This phenomenon confirms that the proposed strategy has a better optimization ability in the training process and exhibits more favorable performance. In Figure \ref{fig3}(b), the lateral position, which indicates lateral displacements of the ego vehicle between the initial and termination of the episode, can directly link to the target of lane-change. These curves exhibit a similar trend to the reward curves, which further verifies the advantage of the proposed strategy. The safety of the ego vehicle is demonstrated in Figure \ref{fig3}(c). The proposed strategy produces less than 20\% of risky episodes in the end, compared to 35\% collisions from the baseline strategy. It should be noted that the proposed strategy's asymptotic collision rate does not approach zero due to the cutoff exploration rate (see Table \ref{table2}). Overall, the proposed strategy offers a higher level of safety during training thanks to the safe demonstration mechanism.

\begin{table}[htbp]
\caption{Performance of decision-making strategies}
\begin{center}
\begin{tabular}{  p{1.8cm} p{1.4cm} p{1.4cm} p{2.4cm}   } 
 \hline
 \textbf{Stratgy} & \textbf{Collision rate (\%), \(\downarrow\)} & \textbf{Success rate (\%), \(\uparrow\)} & \textbf{Average velocity of traffic flow (m/s), \(\uparrow\)} \\ 
 \hline
 Proposed & \textbf{0.0} & \textbf{80.0} & \textbf{8.697 \(\pm\) 0.497}\\ 
 Vanilla-D3QN & 16.7 & 70.0 & 8.458 \(\pm\) 0.703 \\ 
 IL & 23.3 & 73.3 & 8.359 \(\pm\) 0.709\\
 \hline

\label{table3}
\end{tabular}
 \begin{tablenotes}
 \small
 \item Note: the traffic flow refers to all the surrounding vehicles in the scenario. Results are collected from 30 runs with different random seeds.
 \end{tablenotes}
\end{center}
\end{table}

The decision-making abilities of the well-trained strategies were then investigated. Involved candidates include the proposed, vanilla-D3QN-based and IL-based strategies, and each strategy was run 30 times with different random seeds. 

For an overall evaluation, collision rate, success rate of the ego vehicle, and average velocity of the traffic flow were calculated among three strategies, as shown in Table \ref{table3}. An episode without collision is considered safe. The proposed strategy ensures safe driving throughout the test, while the collision events occur in vanilla D3QN and IL. This superiority demonstrates the effect of safe demonstrations. A successful episode means the ego vehicle achieves the objective, namely changing to the shoulder lanes, and therefore the success rate is an indication of the strategy's ability to accomplish the objective. The proposed strategy produces the greatest number of successful episodes. Further, the average velocity of the traffic flow reflects the impact of lane changes on surrounding vehicles. The proposed strategy outperforms the two baseline strategies significantly. This can be due to the better safety in lane changes: the ego vehicle lessens its impact on the rear vehicle, which does not have to undergo intense decelerations. Overall, the proposed strategy exhibits advantageous safety and performance in making lane-change decisions in the studied scenario. 

\begin{figure}[htp]
    \centering
    \includegraphics[width=\linewidth]{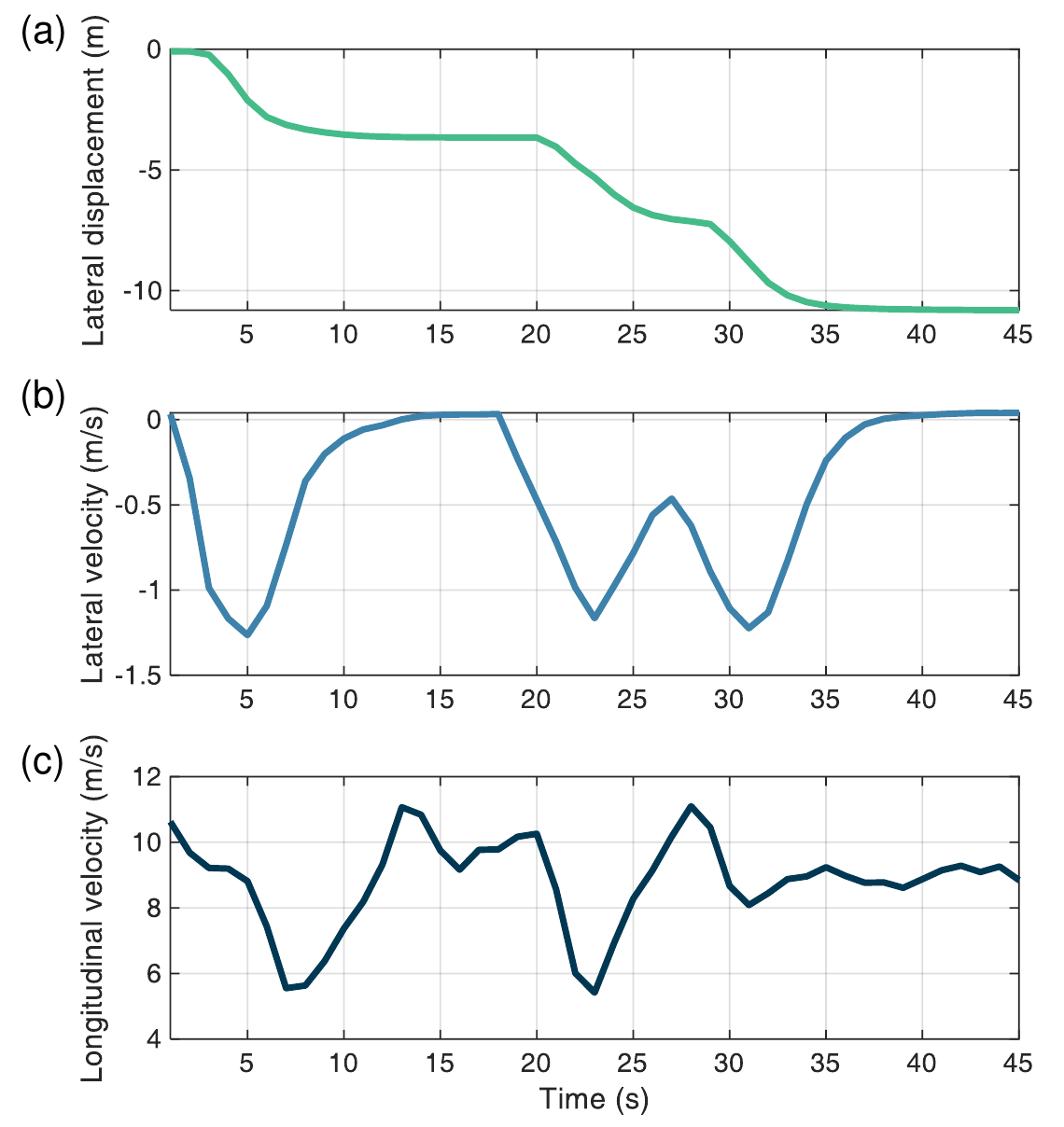}
    \caption{Testing metrics led by the proposed strategy in one episode: (a) lateral displacement of the ego vehicle, (b) lateral velocity of the ego vehicle, and (c) longitudinal velocity of the ego vehicle.}
    \label{fig4}
\end{figure}

A detailed analysis of an episode led by the proposed strategy was then conducted. Lateral displacement and velocity, and longitudinal velocity of the ego vehicle were recorded and illustrated in Figure \ref{fig4}. A series of smooth lane-change was witnessed: the lateral velocity was limited to less than 1.3 m/s. It was also found that although brakes were used during the lane-change period, longitudinal velocities did not vary greatly, which ensures a stable driving process.

In summary, owing to safe demonstrations from human, the proposed strategy achieved an advantageous training performance over the vanilla D3QN, as well as superior decision-making ability in the testing stage over both vanilla D3QN and IL. The effectiveness of the proposed strategy is therefore validated.

\section{Conclusion}\label{section5}
To tackle the safe lane-change problem of AVs, a decision-making strategy based on a human-demonstration-aided RL method is proposed in this study. Taking the D3QN-RL algorithm as the backbone, safe demonstrations, which are generated by a human participant, are utilized to improve the performance of RL on smart decision-making. Our strategy is compared with a vanilla D3QN-based strategy that does not utilize safe demonstrations and an IL-based strategy. Results show that: 

1) the proposed strategy achieves superior safety when compared to existing ones; 

2) the proposed strategy exhibits a better target achievement ability and makes less impact on the surrounding traffic. 

For future work, more effort is needed from different research domains to further advance decision-making and autonomous driving technologies\cite{he_tiv,hu_tiv,hu_vtm,ais,hao_tvt,yiran_ais}.

%\addtolength{\textheight}{-12cm}   % This command serves to balance the column lengths
                                  % on the last page of the document manually. It shortens
                                  % the textheight of the last page by a suitable amount.
                                  % This command does not take effect until the next page
                                  % so it should come on the page before the last. Make
                                  % sure that you do not shorten the textheight too much.

%%%%%%%%%%%%%%%%%%%%%%%%%%%%%%%%%%%%%%%%%%%%%%%%%%%%%%%%%%%%%%%%%%%%%%%%%%%%%%%%

%%%%%%%%%%%%%%%%%%%%%%%%%%%%%%%%%%%%%%%%%%%%%%%%%%%%%%%%%%%%%%%%%%%%%%%%%%%%%%%%

%%%%%%%%%%%%%%%%%%%%%%%%%%%%%%%%%%%%%%%%%%%%%%%%%%%%%%%%%%%%%%%%%%%%%%%%%%%%%%%%
%\section*{APPENDIX}

%Appendixes should appear before the acknowledgment.

\section*{ACKNOWLEDGMENT}

This work was supported in part by A*STAR Grant (No. W1925d0046), A*STAR AME Young Individual Research Grant (No. A2084c0156), the SUG-NAP Grant of Nanyang Technological University, Singapore, and the Urban Mobility Grand Challenge Fund by Land Transport Authority of Singapore (No. UMGC-L010).

\bibliographystyle{ieeetr}
\bibliography{itsc.bib}

\begin{thebibliography}{10}

\bibitem{yang_jas}
Y.~Xing, C.~Lv, L.~Chen, and et~al, ``Advances in vision-based lane detection:
  Algorithms, integration, assessment, and perspectives on acp-based parallel
  vision,'' {\em IEEE/CAA Journal of Automatica Sinica}, vol.~5, no.~3,
  pp.~645--661, 2018.

\bibitem{cognitive_design}
C.~K. Allison and N.~A. Stanton, ``Constraining design: applying the insights
  of cognitive work analysis to the design of novel in-car interfaces to
  support eco-driving,'' {\em Automotive Innovation}, vol.~3, no.~1,
  pp.~30--41, 2020.

\bibitem{cyber_attack}
D.~Zhang, C.~Lv, T.~Yang, and P.~Hang, ``Cyber-attack detection for autonomous
  driving using vehicle dynamic state estimation,'' {\em Automotive
  Innovation}, vol.~4, no.~3, pp.~262--273, 2021.

\bibitem{av_survey}
J.~Li, H.~Cheng, H.~Guo, and S.~Qiu, ``Survey on artificial intelligence for
  vehicles,'' {\em Automotive Innovation}, vol.~1, no.~1, pp.~2--14, 2018.

\bibitem{wu2022uncertainty}
J.~Wu, Z.~Huang, and C.~Lv, ``Uncertainty-aware model-based reinforcement
  learning: Methodology and application in autonomous driving,'' {\em IEEE
  Transactions on Intelligent Vehicles}, pp.~1--10, 2022.

\bibitem{sutton2018reinforcement}
R.~S. Sutton and A.~G. Barto, {\em Reinforcement learning: An introduction}.
\newblock MIT press, 2018.

\bibitem{chen2020conditional}
L.~Chen, X.~Hu, B.~Tang, and Y.~Cheng, ``Conditional dqn-based motion planning
  with fuzzy logic for autonomous driving,'' {\em IEEE Transactions on
  Intelligent Transportation Systems}, 2020.

\bibitem{wang2021interpretable}
H.~Wang, H.~Gao, S.~Yuan, H.~Zhao, K.~Wang, X.~Wang, K.~Li, and D.~Li,
  ``Interpretable decision-making for autonomous vehicles at highway on-ramps
  with latent space reinforcement learning,'' {\em IEEE Transactions on
  Vehicular Technology}, vol.~70, no.~9, pp.~8707--8719, 2021.

\bibitem{friji2020dqn}
H.~Friji, H.~Ghazzai, H.~Besbes, and Y.~Massoud, ``A dqn-based autonomous
  car-following framework using rgb-d frames,'' in {\em 2020 IEEE Global
  Conference on Artificial Intelligence and Internet of Things (GCAIoT)},
  pp.~1--6, IEEE, 2020.

\bibitem{zhang2021tactical}
S.~Zhang, Y.~Wu, H.~Ogai, H.~Inujima, and S.~Tateno, ``Tactical decision-making
  for autonomous driving using dueling double deep q network with double
  attention,'' {\em IEEE Access}, vol.~9, pp.~151983--151992, 2021.

\bibitem{wu2021prioritized}
J.~Wu, Z.~Huang, W.~Huang, and C.~Lv, ``Prioritized experience-based
  reinforcement learning with human guidance for autonomous driving,'' {\em
  IEEE Transactions on Neural Networks and Learning Systems}, pp.~1--15, 2022.

\bibitem{huang2022efficient}
Z.~Huang, J.~Wu, and C.~Lv, ``Efficient deep reinforcement learning with
  imitative expert priors for autonomous driving,'' {\em IEEE Transactions on
  Neural Networks and Learning Systems}, 2022.

\bibitem{rajeswaran2017learning}
A.~Rajeswaran, V.~Kumar, A.~Gupta, G.~Vezzani, J.~Schulman, E.~Todorov, and
  S.~Levine, ``Learning complex dexterous manipulation with deep reinforcement
  learning and demonstrations,'' {\em arXiv preprint arXiv:1709.10087}, 2017.

\bibitem{wang2018intervention}
F.~Wang, B.~Zhou, K.~Chen, T.~Fan, X.~Zhang, J.~Li, H.~Tian, and J.~Pan,
  ``Intervention aided reinforcement learning for safe and practical policy
  optimization in navigation,'' in {\em Conference on Robot Learning},
  pp.~410--421, PMLR, 2018.

\bibitem{wu2020battery}
J.~Wu, Z.~Wei, K.~Liu, Z.~Quan, and Y.~Li, ``Battery-involved energy management
  for hybrid electric bus based on expert-assistance deep deterministic policy
  gradient algorithm,'' {\em IEEE Transactions on Vehicular Technology},
  vol.~69, no.~11, pp.~12786--12796, 2020.

\bibitem{gulcehre2019making}
C.~Gulcehre, T.~Le~Paine, B.~Shahriari, M.~Denil, M.~Hoffman, H.~Soyer,
  R.~Tanburn, S.~Kapturowski, N.~Rabinowitz, D.~Williams, {\em et~al.},
  ``Making efficient use of demonstrations to solve hard exploration
  problems,'' in {\em International conference on learning representations},
  2019.

\bibitem{wang2016dueling}
Z.~Wang, T.~Schaul, M.~Hessel, H.~Hasselt, M.~Lanctot, and N.~Freitas,
  ``Dueling network architectures for deep reinforcement learning,'' in {\em
  International conference on machine learning}, pp.~1995--2003, PMLR, 2016.

\bibitem{wu2021human}
J.~Wu, Z.~Huang, C.~Huang, Z.~Hu, P.~Hang, Y.~Xing, and C.~Lv,
  ``Human-in-the-loop deep reinforcement learning with application to
  autonomous driving,'' {\em arXiv preprint arXiv:2104.07246}, 2021.

\bibitem{dosovitskiy2017carla}
A.~Dosovitskiy, G.~Ros, F.~Codevilla, A.~Lopez, and V.~Koltun, ``Carla: An open
  urban driving simulator,'' in {\em Conference on robot learning}, pp.~1--16,
  PMLR, 2017.

\bibitem{he_tiv}
X.~He, H.~Yang, Z.~Hu, and C.~Lv, ``Robust lane change decision making for
  autonomous vehicles: An observation adversarial reinforcement learning
  approach,'' {\em IEEE Transactions on Intelligent Vehicles}, 2022.

\bibitem{hu_tiv}
Z.~Hu and et~al, ``Driver anomaly quantification for intelligent vehicles: A
  contrastive learning approach with representation clustering,'' {\em IEEE
  Transactions on Intelligent Vehicles}, 2022.

\bibitem{hu_vtm}
Z.~Hu, Y.~Zhang, Y.~Xing, Y.~Zhao, D.~Cao, and C.~Lv, ``Toward human-centered
  automated driving: A novel spatiotemporal vision transformer-enabled head
  tracker,'' {\em IEEE Vehicular Technology Magazine}, pp.~2--9, 2022.

\bibitem{ais}
C.~Lv, Y.~Li, and et~al, ``Human--machine collaboration for automated driving
  using an intelligent two-phase haptic interface,'' {\em Advanced Intelligent
  Systems}, vol.~3, no.~4, p.~2000229, 2021.

\bibitem{hao_tvt}
H.~Chen and et~al, ``Rhonn modelling-enabled nonlinear predictive control for
  lateral dynamics stabilization of an in-wheel motor driven vehicle,'' {\em
  IEEE Transactions on Vehicular Technology}, pp.~1--1, 2022.

\bibitem{yiran_ais}
Y.~Zhang, P.~Hang, C.~Huang, and C.~Lv, ``Human-like interactive behavior
  generation for autonomous vehicles: A bayesian game-theoretic approach with
  turing test,'' {\em Advanced Intelligent Systems}, vol.~4, no.~5, p.~2100211,
  2022.

\end{thebibliography}

\end{document}